\documentclass[a4paper,twoside]{article}

\usepackage{epsfig}
\usepackage{subcaption}
\usepackage{calc}
\usepackage{amssymb}
\usepackage{amstext}
\usepackage{amsmath}
\usepackage{amsthm}
\usepackage{multicol}
\usepackage{makecell}
\usepackage{pslatex}
\usepackage{apalike}
\usepackage{algorithm2e}
\usepackage[bottom]{footmisc}
\usepackage{multirow}
\usepackage{graphicx} % Required for inserting images
\usepackage[nolist]{acronym}
\usepackage{url}

\usepackage{booktabs}    % For professional looking tables
\usepackage{xcolor}      % For coloring rows
\usepackage{array}       % For better column control
\usepackage{geometry}    % For nice margins
\usepackage{tikz}
\definecolor{lightgray}{gray}{0.95}

\usepackage{SCITEPRESS}

\begin{document}

\begin{acronym}
    \acro{mot}[MOT]{Multi-Object Tracking}
    \acro{tbd}[TbD]{Tracking-By-Detection}
    \acro{jde}[JDE]{Joint Detection and Embedding}
    \acro{cnn}[CNN]{Convolutional Neural Network}
    \acro{iou}[IoU]{Intersection-Over-Union}
    \acro{fn}[FN]{False-Negative}
    \acro{fp}[FP]{False-Positives}
    \acro{deta}[DetA]{detection accuracy}
    \acro{assa}[AssA]{association accuracy}
    \acro{pp}[pp]{percentage points}
\end{acronym}

\title{Video-based Locomotion Analysis for Fish Health Monitoring}

\author{\authorname{Timon Palm\sup{1}\orcidAuthor{0009-0003-6016-7718}, Clemens Seibold\sup{1,2}\orcidAuthor{0000-0002-9318-5934}, Anna Hilsmann\sup{1}\orcidAuthor{0000-0002-2086-0951} and Peter Eisert\sup{1,2}\orcidAuthor{0000-0001-8378-4805}}
\affiliation{\sup{1}Fraunhofer HHI, Berlin, Germany}
\affiliation{\sup{2}Humboldt University of Berlin, Berlin, Germany}
% \affiliation{\sup{2}Department of Computing, Main University, MySecondTown, MyCountry}
\email{\{timon.palm, clemens.seibold\, anna.hilsmann, peter.eisert\}@hhi.fraunhofer.de}
}


\abstract{
Monitoring the health conditions of fish is essential, as it enables the early detection of disease, safeguards animal welfare, and contributes to sustainable aquaculture practices. Physiological and pathological conditions of cultivated fish can be inferred by analyzing locomotion activities. In this paper, we present a system that estimates the locomotion activities from videos using multi object tracking. The core of our approach is a YOLOv11 detector embedded in a tracking-by-detection framework. We investigate various configurations of the YOLOv11-architecture as well as extensions that incorporate multiple frames to improve detection accuracy. Our system is evaluated on a manually annotated dataset of \emph{Sulawesi ricefish} recorded in a home-aquarium-like setup, demonstrating its ability to reliably measure swimming direction and speed for fish health monitoring. The dataset will be made publicly available upon publication.
}

% Monitoring the health conditions of fish shoals is central to effective aquaculture management, as swimming behavior is a key indicator of physiological and pathological states. Accurate Multi-Object Tracking is therefore essential. In this work, we demonstrated that increasing the receptive field of the first convolutional layer in YOLOv11 to incorporate larger temporal context windows significantly improves detection performance for Tracking-By-Detection trackers. Even though, the enhancement has only slightly influenced the tracking performance, accurate and consistent quantitative measures of the swimming directions could be derived from all inspected model variations and thus, can be used as a reliable indicator for the fish health conditions.

% Accurate monitoring of fish behavior is critical for sustainable aquaculture, but tracking multiple fish in dynamic environments remains challenging. We introduce a robust multi-fish tracking system and show that the derivation of quantitative directional motion features can reliably used as fish health indicators. Increasing the receptive field of the first convolutional layer in YOLOv11 to incorporate larger temporal context windows improves 

% leads to higher detection accuracy. While the overall tracking precision in Tracking-By-Detection trackers increases only marginally, tracking with single-frame and context-modified models both provide robust and consistent measurements of swimming direction.

\onecolumn \maketitle \normalsize \setcounter{footnote}{0} \vfill

\section{\uppercase{Introduction}}
Image-based fish tracking algorithms play a central role in digital aquaculture, enabling the monitoring of various aspects related to the welfare and reproductive behavior of cultured fish. Accurate tracking of individual fish and entire shoals facilitates detailed behavioral analysis, which is essential for effective aquaculture management.

Quantitative locomotion features derived from trajectories can provide important indicators of fish health. Average speed, burst patterns, acceleration, body orientation, and swimming direction can reveal subclinical stress or pathology. For example, electrical leakage currents, which can arise from malfunctioning equipment, can cause neuromuscular injuries, resulting in loss of equilibrium. This can manifest as abnormal swimming behavior with erratic bursts and atypical locomotion in vertical directions~\cite{Moccia91FishElectrocution}. Swimming performance is widely used as a health and welfare parameter in aquaculture~\cite{leatherland2006fish} and helps to detect diseases and disorders at an early, treatable stage. The often-occurring swim bladder disease, for example, can be caused by various factors such as infection, overfeeding, incorrect water temperature, and stress manifests. It usually results in abnormal swimming, causing fish to float up and down instead of swimming horizontally~\cite{roberts2024SwimBladderDisease}.

Vision-based (i.e., image- and video-based) tracking approaches are particularly attractive due to their simple setup, low cost, durability and no interference with the fish, compared to alternative sensing modalities (e.g., acoustic or biosensor-based systems).

\begin{figure}[!ht]
  \vspace{-0.2cm}
  \centering
    {\includegraphics[width=7.5cm]{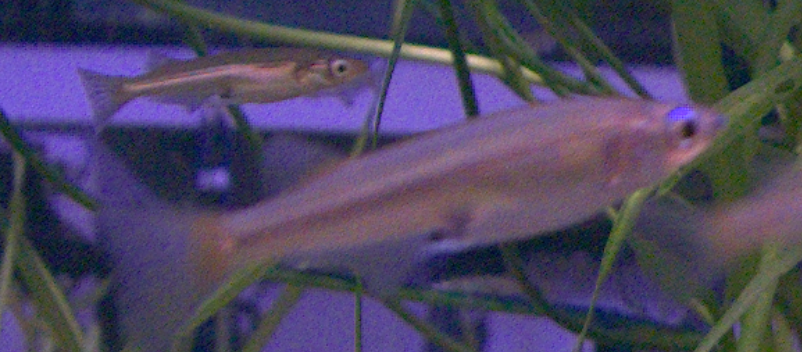}}
  \caption{The \emph{Sulawesi ricefish}: Its lucid appearance, small size and the large number of instances in a shoal makes accurate tracking difficult.}
  \label{fig:ricefish}
\end{figure}

\ac{mot} has been an active research field in computer vision ~\cite{mot:literatureReview} and has been applied to various tracking problems (e.g. cars~\cite{example:vehicle1} or pedestrians~\cite{example:pedestrain1}). However, tracking fish in their natural environment poses significant challenges for vision-based systems, as fish are frequently occluded, interact with one another, change direction rapidly, and undergo continuous morphological deformation due to their non-rigid bodies. Furthermore, unlike humans, fish often exhibit high visual similarity, making individual discrimination more difficult.

However, extracting such behavioral indicators requires consistent long-term detection and tracking of individual fish in complex environments.

In this work, we build upon the classical \ac{tbd} paradigm and enhance YOLO-based detectors by increasing the input channels to allow multi-frame input training. Inspired by recent results showing that multi-frame conditioning can substantially improve detection of small or partially occluded pedestrians~\cite{quan2025lightweightmultiframeintegrationrobust}, we investigate whether similar design principles benefit fish tracking in dense aquaculture scenes. Our evaluation focuses on the challenging \emph{Sulawesi ricefish} whose small size and large shoal densities make them a representative benchmark in controlled culture conditions (see Figure~\ref{fig:ricefish}).

\begin{figure}[!h]
    \vspace{-0.2cm}
    \centering
    {\includegraphics[width=7.5cm]{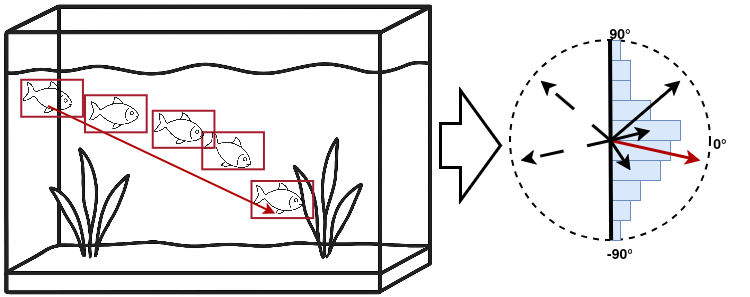}}
    \caption{A schematic view of extracting quantitative directional information of fish observations from Tracking-By-Detection methods.}
    \label{fig:schema}
\end{figure}

Based on the tracking, we derive the distribution of swimming directions for fish shoal in videos, which facilitates the assessment of excessive up and down swimming behavior. This is schematically visualized in Figure~\ref{fig:schema}.

Our contributions are threefold
\begin{itemize}
    \item A curated hand-labeled video dataset of \emph{Sulawesi ricefish} shoals recorded under controlled aquaculture conditions, providing a fish tracking benchmark with complex scenarios, such as rapid direction changes, frequent occlusions, etc. (Dataset:~\url{https://cvg.hhi.fraunhofer.de/})
    \item A qualitative evaluation of multi-frame YOLO adaptations for detecting and tracking fish shoals, examining the impact of incorporating different relative context frames in combination with various YOLO model scales.
    \item A pipeline for the derivation of quantitative locomotion features from fish trajectories as a health indicator.
\end{itemize}

% \begin{itemize}
%     \item multi object tracking
%     \item application of fish tracking (classification, behavior analysis, monitoring well-being, environmental changes)
%     \item motivated the need to monitor the well-being based on the fishs swimming direction
%     \item fish tracking approaches
%     \item highlight the challenges of these particular videos (other blurry, rapid movements, occlusions, number and size of instances, simmilarity)
%     \item explain yolo
%     \item related work: add frames as channel
%     \item summarize main contributions: dataset, larger window size better accuracy (so significant effect on tracking and the assessment of directional movement)
% \end{itemize}
\section{\uppercase{Related Work}}
\label{sec:related-work}

A common class of trackers uses the framework of \ac{tbd}, which splits the task in two stages: first, detecting bounding boxes with Deep Learning models and then using an autoregressive filter (e.g., Kalman filter~\cite{kalman}) to propagate instances of detected bounding boxes over time. Other types of Deep Learning trackers, such as \ac{jde}, jointly learn the detection and tracking phases by extracting object specific embeddings. \ac{jde}s show strong performance for tracking crowds or vehicles, but generally fail to track fish due to their frequent occlusion and inconsistent scales and shapes~\cite{CMFTNetMultipleFish2022}.

Bytetrack~\cite{Zhang_Sun_Jiang_Yu_Weng_Yuan_Luo_Liu_Wang_2022} and BoT-SORT~\cite{BoTSORTRobustAssociations2022} are two simple yet widely used \ac{tbd} methods. Both methods use the Kalman filter~\cite{kalman} as their core motion model. While previous methods~\cite{Wang_2018,du2022giaotrackercomprehensiveframeworkmcmot} incorporated additional appearance feature in the data association, Bytetrack and BoT-SORT rely solely on high-performing detection models and their detection scores. High confidence detections are prioritized in the matching process, while low scoring detections are not entirely discarded but instead considered in a secondary association stage.

The success of detection-only trackers is crucially dependent on the performance of the detection model. The YOLO architecture~\cite{redmon2016lookonceunifiedrealtime} provides a powerful yet lightweight detection model for real-time trackers. It shows superior performance compared to other models such as Faster R-CNN~\cite{LIU2024FishTrack}. Further advancements include YOLO tracking endowed with a CNN for modeling the appearance features of fish~\cite{wojke2017DeepSORT}, its successor incorporates a ResNet-50 backbone~\cite{du2023strongsortmakedeepsortgreat}.

\cite{quan2025lightweightmultiframeintegrationrobust} showed that by minimally modifying the YOLOv7 detection model to accommodate multi-frame inputs, the accuracy and robustness of the detections could be improved up to 8\ac{pp}, in particular for lightweight-model versions.  Formally, given a sequence of $T$ RGB video frames $\{I_t\}^T_t$ with $I_t \in \mathbb{R}^{N\times M \times 3}$, instead of predicting detections for frame $I_t$ only on its information, you concatenate $n$ consecutive frames channel-wise and thereby provide the model with a short time-window of frames $\{I_{t-n+1},\dots, I_t\}$, resulting in the input shape $(N, M, 3\times n)$. Standard convolution is jointly applied across all channels, enabling the integration of spatial-temporal features from the very first layer.

Larger steps in between frames broaden the temporal context available for the model without engorging the channels. This has been shown to be crucial for maximizing the robustness using three to seven frames $n \in \{3, ,7\}$ as context window size \cite{quan2025lightweightmultiframeintegrationrobust}. 

However, the multi-frame YOLO model used for tracking has thus far only been evaluated on pedestrian-tracking tasks. Its capability to handle complex multi-fish tracking scenarios involving frequent occlusions and pronounced morphological variations remains unknown.

The exhaustive study on imaging technologies for fish behavior analysis by ~\cite{cui_fish_2025} has revealed that Bytetrack and BoT-SORT are particularly suitable for the application of fish tracking in different scenarios. However, they further emphasize that comprehensive and unified tracking performance assessment is limited due to a paucity of appropriate datasets. The available datasets either have very low resolution or consist of videos recorded in simplified environments, such as observation tanks with shallow water, uniform backgrounds, and no aquatic plants.
\section{\uppercase{Method}}
\label{sec:method}

The extraction of quantitative directional properties of a fish shoals motion with \ac{tbd} trackers takes three stages: mere detection of instances from imagery, tracking the detection across frames and the computation of directional motion. A schematic illustration of the procedure is shown in Figure~\ref{fig:schema}.

Accurate fish detection from single imagery has its limitations, where fish occlude each other, are blurry due to rapid motion or are poorly illuminated. Intuitively, detection models should benefit from extended temporal context, which enables them to exploit inter-frame differences and fish-specific motion patterns as cues that distinguish fish from other objects.

YOLOv11~\cite{khanam2024yolov11} is a fast and flexible object detection model that predicts object classes and locations in a single pass through the full image, representing a significant advancement in real-time detection performance. The model comprises a \ac{cnn}-based feature extraction backbone and multiple task-specific heads for detection and classification. Extending the channel dimension of the first convolutional layer to accommodate multi-frame input represents a minimal architectural modification that introduces only a negligible increase in parameters. Nevertheless, this adjustment allows the model to exploit temporal information already in the early stages of feature extraction.

In the following, we denote the focus frame, for which the fish detection is performed, as \textbf{X}. Additional frames included in the temporal context are denoted as \textbf{x}, whereas \textbf{\_} indicates frames that are omitted. For example, the context scheme \textbf{x\_X\_x} represents a window size of three frames, where one preceding and one subsequent frame are included, while the frames immediately adjacent to the focus frame are skipped.

The detection models are applied within the Bytetrack and BoT-SORT framework without further modifications, to obtain consistent identity tracks. \ac{tbd} trackers have shown to be powerful, yet lightweight enough for precise tracking in moderate conditions, provided with a sufficiently accurate detection model.

Building on fish trajectories, we further analyze the motion characteristics of the detected individuals by estimating their swimming directions.

We create a direction estimate for each individual fish by gathering the displacements per frame. Thereby, we obtain $n-1$ directions, including the swimming magnitude. Furthermore, we mirror all directions on the vertical axis to one side to eliminate the horizontal directions and to ultimately obtain the angular direction in a degree range of $-90^\circ$ (orthogonal downwards) and $90^\circ$ (orthogonal upwards). 

For noise-reduction purpose, we discarded all trajectories shorter than 5 frames and took the mean direction over 5 frames.

For each video, we obtain a distribution of swimming angles across all fish, which can reveal abnormalities in the collective swimming behavior of the shoal. Under normal conditions, the average swimming direction is predominantly horizontal, and the angle distribution should therefore peak at $0^\circ$ and gradually taper toward both sides. Pronounced peaks away from the center indicate substantial upward or downward motion.

\section{\uppercase{Dataset}}

%To facilitate automated analysis of \emph{Sulawesi ricefish} behavior, we curated a novel tracking and segmentation dataset consisting of high-resolution video recordings under controlled laboratory conditions. The dataset comprises two video sequences (a total of 202 frames) recorded from side and front views at 30 fps and a resolution of $2448\times2048$\,px using an external camera. The recordings were conducted under uniform lighting conditions in an observation tank containing aquatic plants (see Figure~\ref{fig:datasetexample}).

% \begin{figure*}[!htb]
%   \vspace{-0.2cm}
%   \centering
%     {\includegraphics[width=\textwidth]{images/examples_high_contrast.png}}
%   \caption{Example frames with segmentation annotation from our \emph{Sulawesi ricefish} dataset.}
%   \label{fig:datasetexample}
% \end{figure*}

\begin{figure*}[htbp]
    \vspace{-0.2cm}
    \centering
    \begin{minipage}{\columnwidth}
        \centering
        \includegraphics[width=0.7\columnwidth]{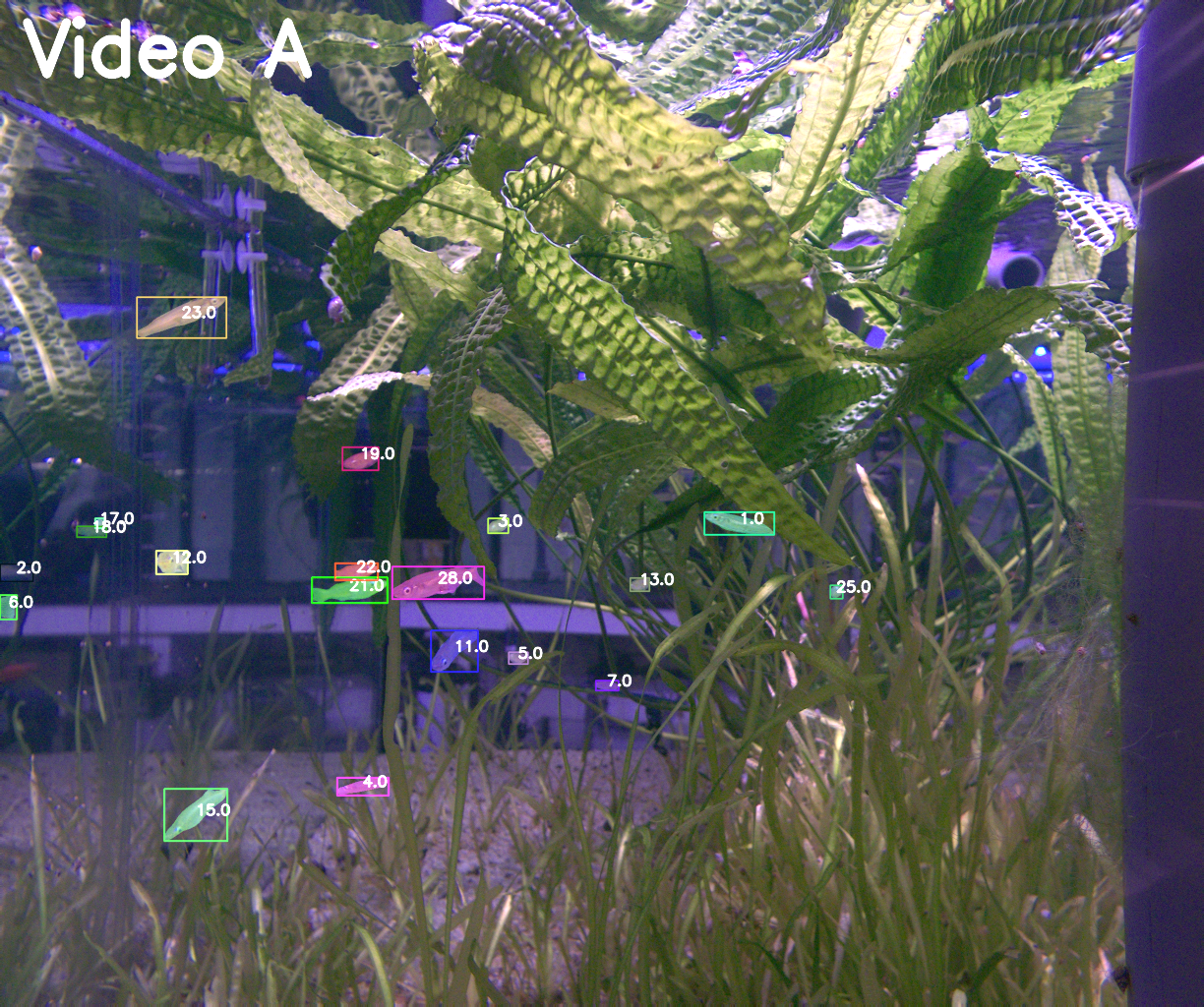}
    \end{minipage}%
    \begin{minipage}{\columnwidth}
        \centering
        \includegraphics[width=0.7\columnwidth]{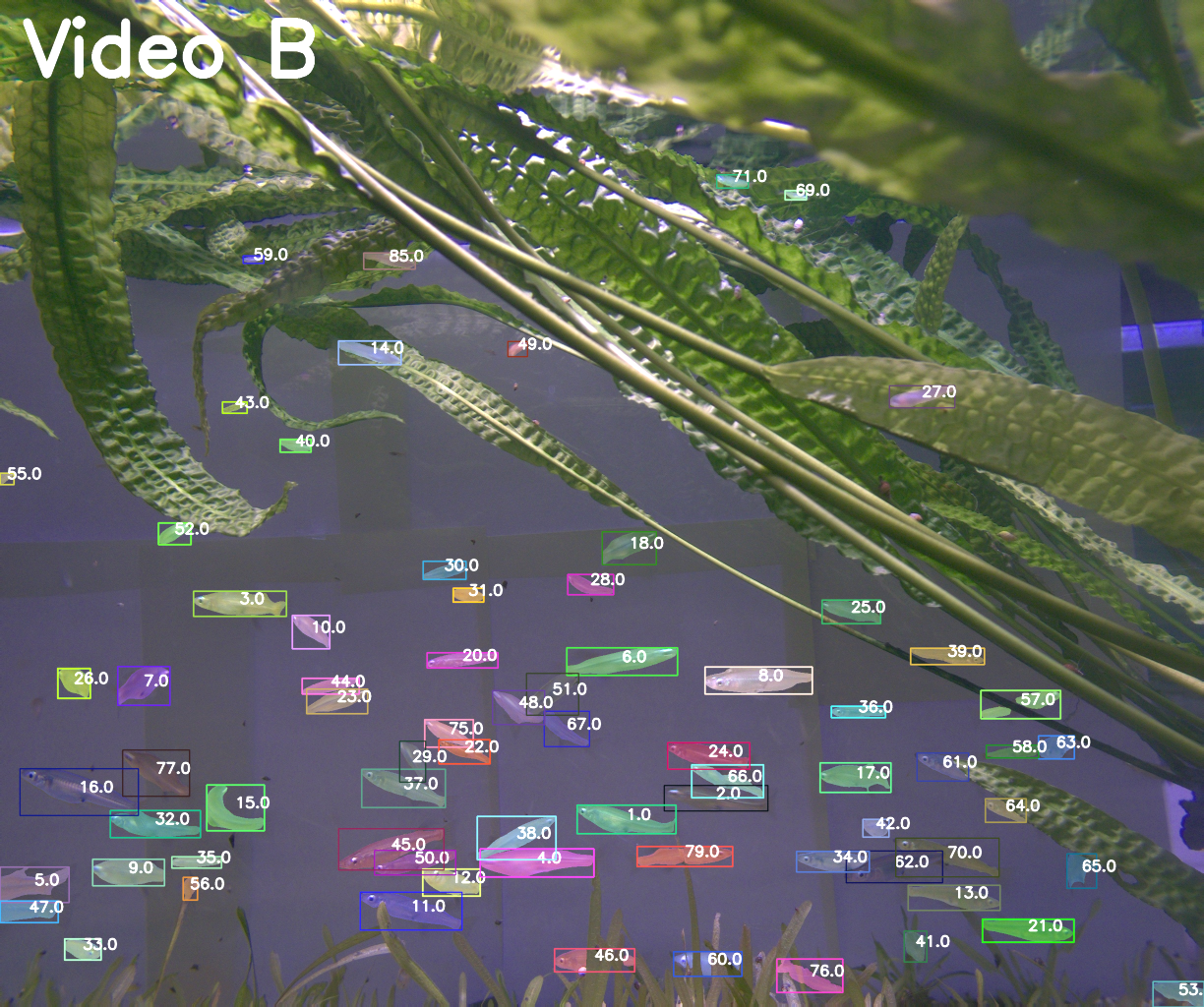}
    \end{minipage}
    \caption{Example frames with segmentation annotation from our \emph{Sulawesi ricefish} dataset.}
    \label{fig:datasetexample}
\end{figure*}

To facilitate automated analysis of Sulawesi ricefish behavior, we curated a novel dataset for tracking and segmentation. We recorded videos of a fish tank at an institution focused on preserving maritime biodiversity through cultivation of endangered fish species. The videos were captured with a Basler a2A2448-75ucPRO and a 6mm lens from multiple viewpoints around the tank, with the camera perpendicular targeting glass wall. The videos have a resolution of 2448 x 2048 pixels at 30 frames per second. The environment resembles a typical home aquarium, with the main light source above the tank. 

The dataset contains two annotated video sequences for evaluation (202\,frames in total). For training, we have annotated 37 additional frames, picked at random from different video sequences to provide frames with fish positions independent of each other. Figure~\ref{fig:datasetexample} shows two example frames with annotations from the dataset.

% We annotated two video sequences for evaluation (202\,frames in total) and additional arbitrary frames from other videos of the same setup for training (37\,frames in total). 

Each frame was semi-automatically annotated to generate pixel-level segmentation masks of individual fish and identity-preserving tracking labels across frames, thereby making the dataset suitable for both segmentation and tracking tasks. It captures a wide range of postures, swimming speeds, and inter-individual occlusions, providing a valuable benchmark for multi-object tracking and instance segmentation algorithms in aquatic environments (see Table~\ref{tab:dataset-spec} for a summary).

\begin{table}[htbp]
    \caption{A summary of the \emph{Sulawesi ricefish} dataset specification for video A and B}
    \label{tab:dataset-spec}
    \centering
    
    % Adjust spacing for aesthetics
    \renewcommand{\arraystretch}{0.5}
    \setlength{\tabcolsep}{4pt}
    
    \begin{tabular}{>{\bfseries}l c c c c c}
        \toprule
        \textbf{Video} & \textbf{Fish Count} & \textbf{\#Tracks} & \textbf{\makecell{Fish per \\Frame (avg)}} \\
        \midrule
        A     & 9214  & 256 & $\sim91.2$\\
        B     & 1969  & 45  & $\sim19.5$\\
        \bottomrule
    \end{tabular}
    
\end{table}

We encourage its use in developing and evaluating computer vision models for fish behavior analysis.
\section{\uppercase{Experiments}}
\label{sec:experiments}

\begin{table*}[htbp]
    \caption{Precision, Recall, mAP50~\cite{everingham2015mAP50} and mAP50-95~\cite{lin2015mAP50-95} metrics for each context configuration in our experiments. Best values are shown in bold, second best are underlined.}
    \label{tab:detection-results}
    \centering
    
    \renewcommand{\arraystretch}{0.5}
    \setlength{\tabcolsep}{5pt}
    
    \begin{tabular}{>{\bfseries}l l | c c c c}
        \toprule
        \textbf{Scheme} & \textbf{Model Size} & \textbf{Precision$\uparrow$} & \textbf{Recall$\uparrow$} & \textbf{mAP50$\uparrow$} & \textbf{mAP50-95$\uparrow$} \\
        \midrule
        \multirow{3}{*}{SingleFrame} 
            & nano   & 0.840 & 0.850 & 0.885 & 0.580\\
            & medium & 0.850 & 0.860 & 0.890 & 0.590\\
            & large  & 0.851 & 0.849 & 0.892 & 0.595\\
        \midrule
        \multirow{3}{*}{xxX} 
            & nano   & 0.858 & 0.750 & 0.830 & 0.530\\
            & medium & 0.884 & 0.773 & 0.844 & 0.563\\
            & large  & 0.913 & 0.733 & 0.845 & 0.557\\
        \midrule
        \multirow{3}{*}{xXx} 
            & nano   & 0.895 & 0.815 & 0.887 & 0.548\\
            & medium & \underline{0.941} & 0.802 & 0.912 & 0.588\\
            & large  & 0.855 & 0.782 & 0.869 & 0.547\\
        \midrule
        \multirow{3}{*}{xxXxx} 
            & nano   & 0.892 & 0.844 & 0.911 & 0.608\\
            & medium & 0.899 & 0.857 & \underline{0.928} & 0.628\\
            & large  & 0.828 & 0.872 & 0.910 & 0.614\\
        \midrule
        \multirow{3}{*}{x\_X\_x} 
            & nano   & 0.881 & 0.820 & 0.882 & 0.601\\
            & medium & 0.882 & \underline{0.878} & 0.920 & \underline{0.645}\\
            & large  & 0.933 & 0.805 & 0.896 & 0.588\\
            % dashed line across columns 3–10
            \multicolumn{6}{l}{\makebox[0pt][l]{\tikz{\draw[dashed] (0,0) -- (11,0);}}}\\ %<= hardcoded width
            x\_X\_x\_pt & pre-trained & \textbf{0.983} & \textbf{0.879} & \textbf{0.963} & \textbf{0.766}\\
        \bottomrule
    \end{tabular}
\end{table*}

We conducted a series of experiments on our \emph{Sulawesi ricefish} dataset to evaluate the tracking accuracy of ByteTrack and BoT-SORT, using context-modified detection models of varying sizes. From the resulting trajectories, swimming direction and speed are derived as potential indicators of the fishes’ health condition. Since both ByteTrack and BoT-SORT operate on bounding boxes, we use the smallest enclosing rectangles of the annotated segmentation masks provided in the dataset.

In Section~\ref{sec:exp-detection}, we examine the impact of using multiple input frames on the detection results by fine-tuning the YOLOv11 model~\cite{khanam2024yolov11}. The best-performing detection models are subsequently employed in Section~\ref{sec:exp-tracking} within ByteTrack and BoT-SORT to evaluate the tracking accuracy and to assess their suitability for estimating the speed and directional dynamics of fish swimming behavior in Section~\ref{sec:exp-swim}.

\subsection{Detection}
\label{sec:exp-detection}

To evaluate the effect of increasing the temporal context of the YOLOv11 detection, we have considered context window sizes of three and five frames, as well as the effect of omitting frames that are directly adjacent to the center frame. We primarily focus on symmetric temporal windows, as these help constrain the space of plausible bounding box predictions, thereby supporting more stable model learning. We also include an asymmetric window configuration as a reference. An overview of the context window configurations used in this study can be taken from Table~\ref{tab:detection-results}.

The weights of the convolutional filters of the additional channels' are randomly initialized with values $w \sim \mathcal{N}(0,0.01)$. The model should still primarily use the information provided by the frame to get the detections for, and only take advantage of features improving its detections from surrounding frames. We evaluated the modification on the YOLOv11 nano (2.6\,M parameters), medium (9.4\,M parameters), and large (20.1\,M parameters) variant, all of which employ a $3\times3$ convolutional kernel in the first layer.

The training set of the fish recordings was split into 33 training frames and 4 validation frames. We used the recommended default training parameters and setting from \cite{yolo11_ultralytics} with a modified box loss factor of $box = 10$ to emphasize the training on accurate detections rather than the class accuracy, as we only have one in this case. Training was conducted for 300\,epochs on a single \emph{NVIDIA RTX 6000} GPU and required approximately 10\,minutes for the largest model.

\subsubsection{Metrics}
\label{sec:det-metrics}

Assessing the performance of object detections to ground-truth data requires the assignment of each proposal bounding box to a true detection. This optimal bipartite assignment problem is solved by the Hungarian algorithm with \ac{iou} values as weights and a threshold of $\text{iou} > 0.5$. Unassigned ground-truth bounding boxes count as \ac{fn}, whereas unassigned proposals count as \ac{fp}.

Precision measures the proportion of correctly detected bounding boxes among all predicted boxes, whereas recall quantifies the proportion of true fish instances successfully detected in the scene. While these provide fundamental quantitative indicators, we further assess detection accuracy using the mAP50 and mAP50–95 metrics, which integrate precision and recall as average precision under \ac{iou} thresholds of 0.5 and from 0.5 to 0.95, respectively. The mAP50–95 metric serves as a stricter measure, reflecting both the localization and size accuracy of the predicted bounding boxes.

\subsubsection{Results}

Table~\ref{tab:detection-results} summarizes the detection results. Overall, larger temporal windows tend to yield better performance (best and second-best results are shown in bold and underlined, respectively). Interestingly, the model x\_X\_x with omitted frames performs on par to the full context window model xxXxx, suggesting that temporal sparsity does not necessarily harm detection accuracy. While individual configurations show notable gains, e.g. x\_X\_x model m (up to 5.5\,\ac{pp} better than single frame in mAP50-95), no universally consistent trend emerges across all architectures. The temporal context x\_X\_x for example, shows great improvement in composition with the YOLO model in size medium, whereas it provides little to no improvement for the large model. Generally, the model size medium performed best throughout all context schemes.

The reference with asymmetric context (xxX) performed worse in all experiments, even worse than the single-frame baseline. This confirms the assumption that symmetric context windows are beneficial, whereas context windows that extend only into the past are, in fact, detrimental.

%The initial expectation of using symmetric windows turned out to be beneficial, as the reference window with asymmetric context (xxX) even performed worse than the single frame baseline.

Additionally, to further examine the potential improvement in detection accuracy, we initialized the best-performing context-type model (x\_X\_x) using the weights of our reference single-frame model rather than the default YOLOv11 weights. This variant is denoted as x\_X\_x\_pt in Table~\ref{tab:detection-results}. Taking the pre-trained weights on our fish dataset had an impressive effect on the detection precision, yielding gains of 5\,\ac{pp} over the x\_X\_x configuration and 13\,\ac{pp} over the single frame baseline, and resulting in the highest mAP50-95 score of 76.6\,\% in our experiments. Interestingly, although detection accuracy increased, recall only improved marginally (1-2\,\ac{pp} improvement over baseline). It is important to note, however, that pre-training the single frame model and later using those weights to increase the context size, subjects the model to two optimization phases on the same dataset. As a result, this configuration undergoes roughly twice as many optimization steps as the other models in the comparison.

For evaluating the tracking with Bytetrack and BoT-SORT, we only considered the models x\_X\_x, x\_X\_x\_pt, xXx, as well as the single frame reference model. Again, we conducted experiments on all three YOLOv11 sizes.

\subsection{Tracking}
\label{sec:exp-tracking}

Improving the accuracy of bounding box detection models for \ac{tbd} tracker does not necessarily improve the tracking itself. With this experiment, we examine how much \ac{tbd} trackers can benefit from our improved detection models for fish tracking. We have used the two very commonly applied \ac{tbd} methods Bytetrack~\cite{Zhang_Sun_Jiang_Yu_Weng_Yuan_Luo_Liu_Wang_2022} and BoT-SORT~\cite{BoTSORTRobustAssociations2022} on our \emph{Sulawesi ricefish} dataset, containing two videos. We used the detection models x\_X\_x, x\_X\_x\_pt, xXx and the single frame reference model. Additionally, we ran both tracking methods on the ground-truth detections to set an upper limit for the tracking performance. The default parameters from Ultralytics~\cite{yolo11_ultralytics} are taken for Bytetrack and BoT-SORT.

\begin{table*}[t]
    \caption{Bytetrack: Tracking results for using different context windows for the YOLOv11 detection model. The best performance is highlighted in bold and the second best is underlined.}
    \label{tab:tracking-byte-results}
    \centering
    
    \renewcommand{\arraystretch}{0.5}
    \setlength{\tabcolsep}{4pt}
    
    \begin{tabular}{>{\bfseries}l l | c c c c | c c c c}
        \toprule
        \multirow{2}{*}{\textbf{Scheme}} & \multirow{2}{*}{\textbf{Size}} & \multicolumn{4}{c}{\textbf{Video A}} & \multicolumn{4}{c}{\textbf{Video B}} \\
        & & mAP50-95$\uparrow$ & IDF1$\uparrow$ & MOTA$\uparrow$ & HOTA$\uparrow$ & mAP50-95$\uparrow$ & IDF1$\uparrow$ & MOTA$\uparrow$ & HOTA$\uparrow$ \\
        \midrule
        GT Detections & - & 1.000 & 0.756 & 0.991 & 0.803 & 1.000 & 0.815 & 0.981 & 0.856\\
        \midrule
        \multirow{3}{*}{Single Frame} 
            & nano   & 0.451 & 0.396 & 0.165 & 0.330 & 0.368 & \underline{0.486} & 0.360 & 0.373\\
            & medium & 0.413 & 0.465 & 0.005 & 0.374 & 0.442 & 0.461 & 0.265 & 0.362\\
            & large  & 0.416 & 0.439 & 0.012 & 0.361 & 0.370 & 0.427 & 0.193 & 0.339\\
        \midrule
        \multirow{3}{*}{xXx} 
            & nano   & 0.445 & 0.539 & 0.331 & 0.408 & 0.398 & 0.463 & 0.363 & 0.346\\
            & medium & \textbf{0.502} & 0.559 & 0.368 & \underline{0.428} & 0.420 & 0.451 & 0.339 & 0.341\\
            & large  & 0.413 & 0.442 & 0.130 & 0.362 & 0.281 & 0.390 & 0.196 & 0.301\\
        \midrule
        \multirow{3}{*}{x\_X\_x} 
            & nano   & 0.447 & \textbf{0.583} & 0.357 & \textbf{0.433} & 0.406 & 0.432 & 0.327 & 0.334\\
            & medium & \underline{0.493} & 0.554 & \textbf{0.384} & 0.419 & \textbf{0.477} & \textbf{0.492} & \textbf{0.409} & \textbf{0.375}\\
            & large  & 0.420 & 0.485 & 0.112 & 0.390 & 0.374 & 0.426 & 0.284 & 0.329\\
        \midrule
        \multirow{3}{*}{x\_X\_x\_pt} 
            & nano   & 0.404 & 0.405 & 0.106 & 0.336 & 0.431 & 0.482 & 0.382 & \underline{0.374}\\
            & medium & 0.363 & 0.456 & 0.368 & 0.367 & 0.300 & 0.358 & 0.147 & 0.288\\
            & large  & 0.471 & \underline{0.560} & \underline{0.370} & 0.422 & \underline{0.455} & 0.472 & \underline{0.387} & 0.361\\
        \bottomrule
    \end{tabular}
\end{table*}

\vspace{-1pt}

\begin{table*}[t]
    \caption{BoT-SORT: Tracking results for using different context windows for the YOLOv11 detection model. The best performance is highlighted in bold and the second best is underlined.}
    \label{tab:tracking-bots-results}
    \centering
    
    \renewcommand{\arraystretch}{0.5}
    \setlength{\tabcolsep}{4pt}
    
    \begin{tabular}{>{\bfseries}l l | c c c c | c c c c}
        \toprule
        \multirow{2}{*}{\textbf{Scheme}} & \multirow{2}{*}{\textbf{Size}} & \multicolumn{4}{c}{\textbf{Video A}} & \multicolumn{4}{c}{\textbf{Video B}} \\
        & & mAP50-95$\uparrow$ & IDF1$\uparrow$ & MOTA$\uparrow$ & HOTA$\uparrow$ & mAP50-95$\uparrow$ & IDF1$\uparrow$ & MOTA$\uparrow$ & HOTA$\uparrow$ \\
        \midrule
        GT Detections & - & 1.000 & 0.756 & 0.991 & 0.803 & 1.000 & 0.824 & 0.976 & 0.856\\
        \midrule
        \multirow{3}{*}{Single Frame} 
            & nano   & 0.451 & 0.407 & 0.071 & 0.361 & 0.368 & \underline{0.532} & 0.455 & \underline{0.423}\\
            & medium & 0.413 & 0.507 & 0.130 & 0.428 & 0.442 & 0.509 & 0.383 & 0.421\\
            & large  & 0.416 & 0.489 & 0.099 & 0.422 & 0.370 & 0.489 & 0.313 & 0.400\\
        \midrule
        \multirow{3}{*}{xXx} 
            & nano   & 0.445 & 0.604 & 0.363 & 0.481 & 0.398 & 0.493 & 0.427 & 0.386\\
            & medium & \textbf{0.502} & 0.616 & \underline{0.485} & \textbf{0.494} & 0.420 & 0.507 & 0.407 & 0.396 \\
            & large  & 0.413 & 0.517 & 0.250 & 0.432 & 0.281 & 0.440 & 0.266 & 0.340\\
        \midrule
        \multirow{3}{*}{x\_X\_x} 
            & nano   & 0.447 & \textbf{0.625} & 0.479 & 0.483 & 0.406 & 0.482 & 0.392 & 0.390\\
            & medium & \underline{0.493} & 0.606 & \textbf{0.507} & 0.474 & \textbf{0.477} & \textbf{0.534} & \textbf{0.503} & \textbf{0.429}\\
            & large  & 0.420 & 0.538 & 0.253 & 0.435 & 0.374  & 0.460 & 0.345 & 0.378\\
        \midrule
        \multirow{3}{*}{x\_X\_x\_pt} 
            & nano   & 0.404 & 0.470 & 0.031 & 0.397 & 0.431 & 0.520 & 0.463 & 0.420\\
            & medium & 0.363 & 0.486 & 0.072 & 0.410 & 0.300 & 0.395 & 0.212 & 0.320\\
            & large  & 0.471 & \underline{0.623} & 0.474 & \underline{0.491} & \underline{0.455} & 0.516 & \underline{0.464} & 0.408\\
        \bottomrule
    \end{tabular}
\end{table*}

\subsubsection{Metrics}

Based on the detection assignments described in Section~\ref{sec:det-metrics}, we evaluate the quality and consistency of the tracking results using three \ac{mot} metrics, which have been commonly applied in previous work \cite{Zhang_Sun_Jiang_Yu_Weng_Yuan_Luo_Liu_Wang_2022,BoTSORTRobustAssociations2022}. Each metric captures a distinct aspect of tracking performance:
\begin{itemize}
    \item \textbf{IDF1}~\cite{ristani2016performancemeasuresdataset} is defined as the harmonic mean of the identity precision and identity recall. It quantifies the consistency of object identities across frames. Unlike traditional frame-based precision and recall, IDF1 considers temporal consistency, penalizing identity switches and fragmented tracks. High IDF1 values indicate that the tracker maintains stable and consistent identities across frames.
    \item \textbf{MOTA} (Multiple Object Tracking Accuracy) is part of the CLEAR metrics~\cite{bernardin2008CLEAR} measuring the overall tracking accuracy by jointly accounting for missed detections (FN), False-Positives and identity switches. Although MOTA offers a single scalar summarizing overall performance, it is known to be dominated by detection quality and may underrepresent the impact of identity fragmentation ~\cite{Luiten2020HOTA}. Nonetheless, it remains a standard benchmark for evaluating multi-object trackers due to its interpretability and comparability.
    \item \textbf{HOTA} (Higher Order Tracking Accuracy)~\cite{Luiten2020HOTA} provides a balanced and comprehensive measure of accurate detection, association and localization. It is defined as the geometric mean of detection accuracy (\ac{deta}) and association accuracy (\ac{assa}), thereby integrating both the per-frame and temporal aspects of tracking.
\end{itemize}
To further contextualize these tracking metrics, we relate them to mAP50–95, a detection-only performance measure that reflects the average precision across multiple \ac{iou} thresholds.

\subsubsection{Results}

Tables \ref{tab:tracking-byte-results} and \ref{tab:tracking-bots-results} summarize the tracking results obtained using ByteTrack and BoT-SORT, respectively. For video B, the model with the x\_X\_x context window in size medium consistently outperforms all other models across each metrics, regardless of the tracking algorithm. In contrast, the results for video A are less pronounced. While the same model still performs well relative to others, it does not achieve the highest scores on every metric. Interestingly, the nano model variant achieved slightly higher IDF1 and HOTA scores for video A with ByteTrack, though the improvement was modest (1–3\,\ac{pp}).

With the exception of x\_X\_x\_pt, the large model variant consistently performed worse. This is likely attributable to overfitting to the training data, as its detection mAP50-95 on both test videos is markedly lower than that of all other models, including the single-frame reference, despite achieving the highest validation performance during training.

Providing larger context windows x\_X\_x could improve the general detection accuracy (mAP50-95) in the dataset up to 9\,\ac{pp} and therefore, leads to better tracking results, too. Though, the tracking improvements measure by IDF1, MOTA and HOTA are only about 1\,\ac{pp}.

Nevertheless, the tracking results are not consistent among model sizes. Although in general the medium model performed better than small, the large model performed worse, in some instances even worse than the nano single frame reference.

% Compared to tracking using ground-truth detections, a substantial performance gap remains, primarily due to inaccurate or missing detections. Both ByteTrack and BoT-SORT are capable of achieving HOTA scores of up to 85\,\% when provided with ground-truth detections.

A clear performance gap persists relative to tracking with ground-truth detections, mainly driven by inaccurate or missing detections. Under ground-truth detection conditions, ByteTrack and BoT-SORT reach HOTA scores of up to 85\,\%.

\subsection{Swimming Direction}
\label{sec:exp-swim}

In this section, we evaluate the reliability and credibility of the swimming direction estimates produced by \ac{tbd} trackers, with a focus on their suitability for fish health monitoring.

% We create a swimming direction estimate for each individual fish by gathering the displacements per frame. Thereby, we obtain $n-1$ directions, including the swimming magnitude. This results in the precise distribution of swimming directions for the ground-truth tracking. However, the \ac{tbd} in our experiments not only yield long and consistent trajectories. Therefore, to reduce the noise, we discarded all trajectories shorter than 5 frames and we took the mean direction over 5 frames. The results exhibit much less fluctuations, albeit the number of direction vectors is less than the ground-truth. If the tracking is valid, the directional distribution should still be similar to the ground-truth, however, each direction ultimately counts more.

% Furthermore, we mirror all directions on the vertical axis to one side to eliminate the horizontal directions and to ultimately obtain the angular direction in a degree range of $-90^\circ$ (orthogonal downwards) and $90^\circ$ (orthogonal upwards). This process is illustrated in Figure \ref{fig:schema}.

Due to noise-reduction by discarding very short trajectories, the number of resulting direction vectors is lower than in the ground-truth. Nevertheless, if the tracking is accurate, the overall directional distribution should remain similar to the ground-truth, even though each direction effectively contributes a higher weight.

\begin{figure*}[!t]
  \vspace{-0.2cm}
  \centering
    {\includegraphics[width=\textwidth]{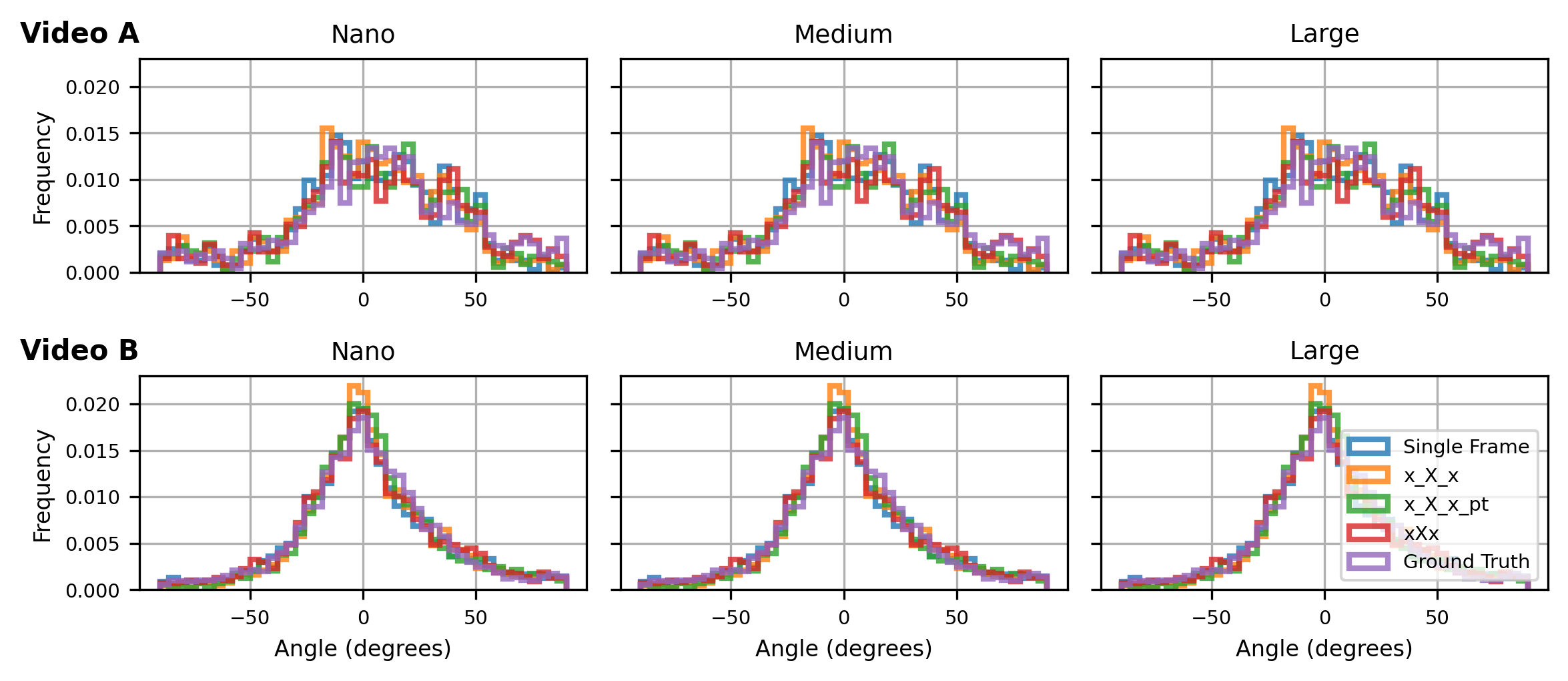}}
  \caption{Histograms depicting the angular swimming directions extracted from tracking with model size nano (left), medium (middle) and large (right) for video A (top row) and video B (bottom row).}
  \label{fig:swimDir}
\end{figure*}

The results of this analysis are depicted in Figure~\ref{fig:swimDir}, showing the swimming direction distribution for video A and B from our \emph{Sulawesi ricefish} dataset. We examined the directions of all context variants from the tracking in Section \ref{sec:exp-tracking} and all three model sizes (nano, medium and large).

\begin{figure*}[!htb]
  \vspace{-0.2cm}
  \centering
    {\includegraphics[width=\textwidth]{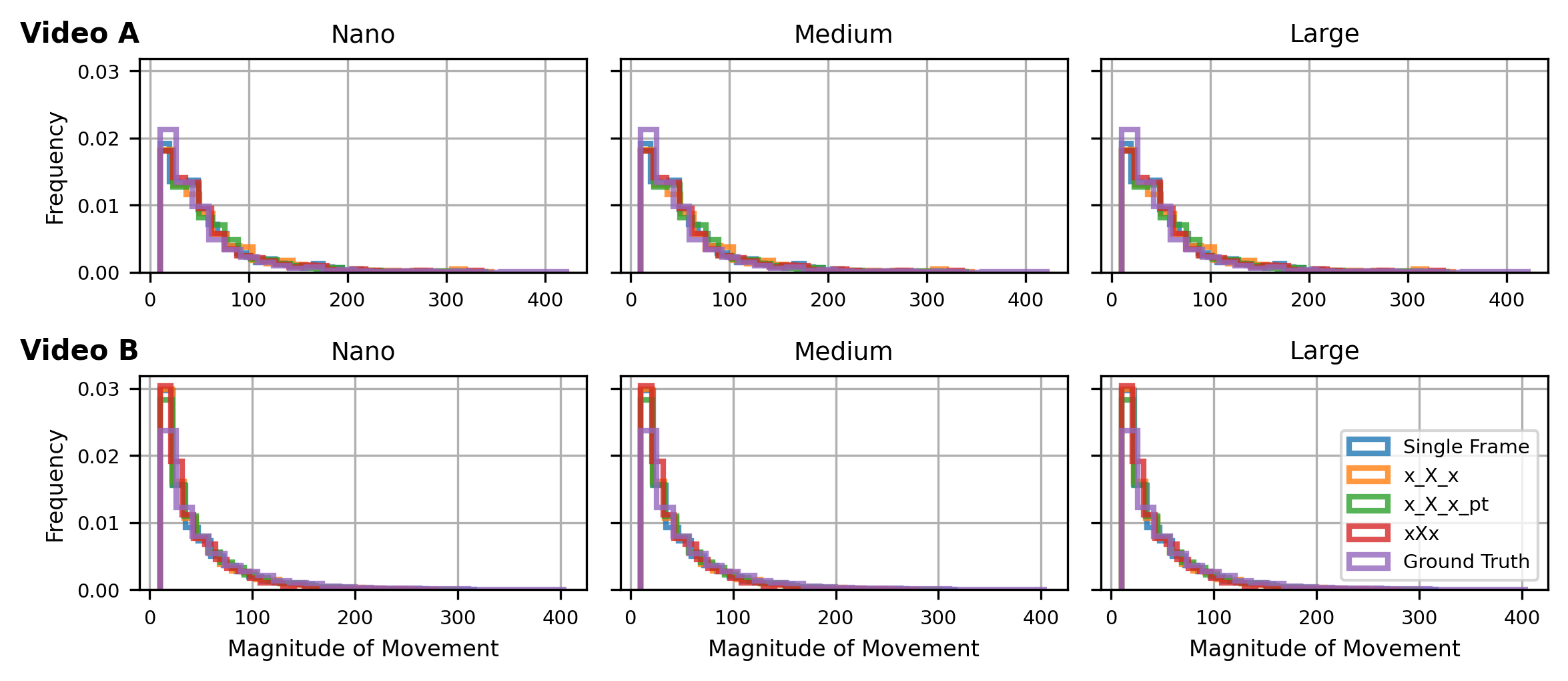}}
  \caption{Histograms showing the aggregated magnitude of the swimming directions from tracking with model size nano (left), medium (middle) and large (right) for video A (top row) and video B (bottom row).}
  \label{fig:magnitudes}
\end{figure*}

Figure~\ref{fig:magnitudes} shows the distribution of the swimming magnitude (speed) in pixel space over 5\,frames. The results are similar to the direction analysis. All model variations are close to the ground-truth, however all variants have larger portion of low magnitude estimates, in particular for video B.

Overall, we can say that the derived directional estimates of the fish motion are sufficiently close to the true directional estimate for all studied model variants, with only a few exceptions (x\_X\_x\_pt in size medium in video A). This is a surprising result: that improving the fish tracking by using larger context windows, the assessment of the swimming directions and magnitude is little affected. Even the classic single frame tracking suffices as a method for computing the directions in a holistic view.

% Overall, nearly all model variants provide direction estimates that are sufficiently accurate for downstream analysis, such as directional assessment,

% Figure \ref{} illustrates the angular deviations of each fish’s trajectory. When viewed as aggregated distributions, positive and negative angular errors may cancel each other out, leading to seemingly credible distributions even though individual angle estimates may deviate substantially from the ground-truth.

\section{\uppercase{Conclusions}}

Quantitative features derived from fish movement can be important indicators for the fish health conditions in cultured environments. In particular, prominent vertical swimming behavior can reveal stress or pathology, potentially caused by e.g. electrical leakage current~\cite{Moccia91FishElectrocution}. To assess the feasibility of inferring fish locomotion activity in aquarium settings, we constructed a \ac{mot} and segmentation dataset comprising shoals of \emph{Sulawesi ricefish}.

We have shown that increasing the context window in YOLOv11 detection models improves detection accuracy with modest adaptations. Higher detection accuracy leads to more robust and consistent tracking in Bytetrack and BoT-SORT tracking methods. Tracking on the ground-truth detections has shown that \ac{tbd} are capable enough for accurate fish tracking. However, the results substantially depend on the fidelity of the detection, which are yet far behind the ground-truth.

For further evaluation, we plan to deploy our tracking system in a biodiversity-focused fish farm and analyze fish locomotion over an extended period of time. In addition, we will extend our tracking approach by estimating locomotion during the camera’s exposure time using concepts for explicit modeling of motion blur~\cite{Seibold2017MotionBlurTracking}.

Nevertheless, current \ac{tbd} methods are sufficient to derive reliable quantitative estimates of shoal motion direction, and these measurements can therefore be used to draw conclusions about the fish’s health status.

\section*{ACKNOWLEDGEMENTS}
This work was partly funded by the German Federal Ministry for Economic Affairs and Energy  (FischFitPro, grant no. 16KN095536) and the Federal Ministry of Research, Technology and Space (REFRAME, grant no. 01IS24073A).

\bibliographystyle{apalike}
{\small
\bibliography{references}}

\end{document}